\documentclass[11pt,a4paper]{article}
\usepackage[hyperref]{acl2020}
\usepackage{times}
\usepackage{latexsym}

\usepackage{color,colortbl}
\usepackage{xcolor}

\usepackage{booktabs}
\usepackage{graphicx}
\usepackage{multirow}
\usepackage{array}
\usepackage{tabularx}
\usepackage{amssymb}
\usepackage{pifont}
\usepackage{tikz}
\usepackage{tcolorbox}

\usepackage{balance}

\newcommand{\PreserveBackslash}[1]{\let\temp=\\#1\let\\=\temp}
\newcolumntype{C}[1]{>{\PreserveBackslash\centering}p{#1}}
\newcolumntype{R}[1]{>{\PreserveBackslash\raggedleft}p{#1}}
\newcolumntype{L}[1]{>{\PreserveBackslash\raggedright}p{#1}}
\newcommand{\greencheck}{{\ding{51}}}
\newcommand{\redcross}{{\ding{55}}}

\definecolor{hl}{HTML}{F5FFFA}

\definecolor{yellow-green}{rgb}{0.6, 0.8, 0.2}
\definecolor{tropicalrainforest}{rgb}{0.0, 0.46, 0.37}
\definecolor{rosevale}{rgb}{0.67, 0.31, 0.32}
\usepackage{microtype}

\aclfinalcopy 

\title{\textsc{TimeDial}: Temporal Commonsense Reasoning in Dialog}

\author{Lianhui Qin$^{\diamondsuit}$$^{\thanks{  ~~Work done during an internship at Google.}}$~~~~~~~Aditya Gupta$^\spadesuit$~~~~~~~Shyam Upadhyay$^\spadesuit$\\

\bf Luheng He$^\spadesuit$~~~~~~~~Yejin Choi$^{\diamondsuit}$~~~~~~~Manaal Faruqui$^\spadesuit$ \\
$^\spadesuit$Google Assistant \\
$^\diamondsuit$Paul G. Allen School of Computer Science \& Engineering, University of Washington \\
\texttt{\{gaditya, shyamupa, luheng, mfaruqui\}@google.com} \\
\texttt{\{lianhuiq, yejin\}@cs.washington.edu} \\
  }

\date{}

\begin{document}
\maketitle
\begin{abstract}
Everyday conversations require understanding everyday events, which in turn, requires understanding temporal commonsense concepts interwoven with those events.
Despite recent progress with massive pre-trained language models (LMs) such as \textsc{T5} and \textsc{GPT-3}, 
their capability of temporal reasoning in dialogs remains largely under-explored. 
In this paper, we present the first study to investigate pre-trained LMs for their 
temporal reasoning capabilities in dialogs by introducing a new task and a crowd-sourced English challenge set, \textsc{TimeDial}. We formulate \textsc{TimeDial} as a multiple choice cloze task with over 1.1K carefully curated dialogs. 
Empirical results demonstrate that even the best performing models struggle on this task compared to humans, with $23$ absolute points of gap in accuracy. Furthermore, our analysis reveals that the models fail to reason about dialog context correctly; instead, they rely on shallow cues based on existing temporal patterns in context, 
motivating future research for modeling temporal concepts in text and robust contextual reasoning about them. The dataset is publicly available at: \url{https://github.com/google-research-datasets/timedial}.
\end{abstract}

\section{Introduction}
 \begin{table}[t]
\footnotesize
    \centering
    \begin{tabular}{L{7cm}}
    \toprule
        A: May we see the wine list please.\\
    B: Sure. Our special wine today is a 1989 Chardonnay.\\
    A: \colorbox{hl}{\textbf{I'd like a bottle}} please.\\
    B: I'll need to \colorbox{hl}{\textbf{see your ID}} please.\\
    A: Here you go. \\
    B: Sorry about the inconvenience, you look so young. I had to make sure you are over \rule{1cm}{1pt}.\\
    \\
    a) 21 years old \greencheck \hfill b) 30 years old \redcross\\
    c) 4 years old \redcross \hfill d) 18 years old \greencheck \\
    
    \midrule
    A: ~Good morning! May I help you?\\
    B: ~Yes. My wife and I are interested in renting a house for the summer. \\
    A: ~Very well. How long do you want the house? All summer?\\
    B: ~No, not all summer. Just for \colorbox{hl}{\textbf{six weeks}}.\\
    A: ~I am afraid I can only rent it for \colorbox{hl}{\textbf{two months}}. \\
    B: ~My holiday is only \rule{1cm}{1pt}, but I think my brother  and his family would take it for the other \colorbox{hl}{\textbf{two weeks}}.\\
    \\
    a) six decades \redcross \hfill b) 45 days \greencheck \\
    c) six weeks \greencheck   \hfill d) two months \redcross\\
    \bottomrule
    \end{tabular}
\caption{Examples from our \textsc{TimeDial} challenge set, demonstrating the need for commonsense knowledge and arithmetic reasoning over the context to infer the correct answers. Key contextual information for reasoning success is highlighted.
}  
\label{tab:main_example} 
\vspace{-1em}
\end{table}

Humans can effortlessly reason about temporal concepts of everyday events such as their duration, frequency, or relative ordering~\cite{allen1984towards,radvansky2014event} based on rich commonsense knowledge about how the world works, especially in relation to time. 
However, reasoning about such concepts has been challenging for machines \cite{KAHN197787,kozareva}
 since it requires both understanding the local temporal expressions and reasoning about their global contexts such as their relative ordering and relations \cite{uzzaman-etal-2013-semeval,ning-etal-2018-improving,pustejovsky2017iso}. 
The problem becomes even more challenging in dialogs, where explicit and implicit inter-dependencies among temporal concepts can appear across conversation turns.

For instance, for the first dialog in Table~\ref{tab:main_example}, one must understand the context, i.e., selling wine, and use world knowledge of minimum legal drinking age in order to reason about correct answers to fill in the blank.
Similarly, in the second conversation, commonsense about the durations \emph{summer}, \emph{month}, \emph{week}, \emph{day} and their relations, plus numerical reasoning, are necessary to make the inference.

Although previous works have studied temporal reasoning in natural language, they have either focused on specific time-related concepts in isolation, such as temporal ordering and relation extraction~\cite{leeuwenberg2018temporal,ning-etal-2018-joint}, and/or dealt with limited context, such as single-sentence-based question answering \cite{zhou-etal-2019-going} and natural language inference~\cite{vashishtha-etal-2020-temporal,mostafazadeh-etal-2016-corpus}.

In this work, we make the first systematic study of temporal commonsense reasoning in a multi-turn dialog setting. The task involves complex reasoning that requires operations like comparison and arithmetic reasoning over temporal expressions and the need for commonsense and world knowledge.

We design a new task for dialog-based temporal reasoning and present a new challenge set in English, called \textsc{TimeDial}, to evaluate language understanding models on the task.
We formulate the problem as a crowd-sourced cloze task with multiple choices based on dialogs in the DailyDialog dataset \cite{li2017dailydialog}. Given a dialog with one temporal span masked out, the model is asked to find \emph{all} correct answers from a list of four options to fill in the blank (Table~\ref{tab:main_example}).

The challenge set requires the models to demonstrate understanding of the context and use temporal commonsense to make right choices. 
Our final challenge set consists of $1.1$K carefully curated dialog instances.

We then study the performance of several state-of-the-art pre-trained language models on \textsc{TimeDial} along several dimensions including modeling paradigms (classification, mask filling, and generation), the scope of dialog contexts, in-domain vs. out-of-domain training, dependence on shallow text matching for reasoning, and the types of reasoning required.
Our experiments demonstrate that off-the-shelf, pre-trained language models cannot effectively reason about temporal aspects in a dialog, even with domain-specific finetuning. 
Our findings indicate that large-scale pre-trained models even after fine-tuning may not be sufficient for robust temporal reasoning in dialogs, and motivate future research toward modeling temporal concepts over  diverse everyday events, and contextual reasoning about them. 

\section{Task: Temporal Reasoning in Dialog}
\label{sec:task}

We formulate the dialog-based temporal commonsense reasoning problem as a \textit{cloze} task~\cite{taylor1953cloze}. Formally, given a multi-turn dialog context
of $n$ conversational turns between two speakers A and B, where a temporal words span within the context is masked out, 
the task is to predict the suitable temporal expression(s) for the masked-out span from a list of options. 
That is, we want the conversation model to select all the correct answers from the options based on the dialog context. Following similar cloze-style challenge datasets, we use accuracy as the evaluation metric~\cite{mostafazadeh-etal-2016-corpus,onishi-etal-2016-large,mihaylov2018knowledgeable}.

Having a non-trivial set of options is crucial to build a challenge set and to avoid accidental spurious biases \cite{geirhos2020shortcut, gururangan2018annotation, le2020adversarial}.
We ensure this via the following filtering process. {\bf (1)} For each masked span, there is more than one correct answer in the options. This makes the task more challenging for models since more comprehensive understanding of the context is required to recognize all the correct choices. In our dataset (\S\ref{sec:data}) we guarantee two incorrect answers for each masked span. {\bf (2)} Some incorrect options are selected to be spuriously correlated with the dialog context. For example, we include temporal spans in the dialog context as negative options, which will challenge models that rely primarily only on shallow pattern matching without correct temporal reasoning. We present more information in \S\ref{sec:data} about how the negative options were created by human annotators.

\begin{table*}[!t]
\centering
\footnotesize
\begin{tabular}{L{2cm}L{11cm}L{2cm}}
\bf Category   & \bf Dialog        &  \bf{Options}                            \\
\toprule
\begin{tabular}{@{}L{2cm}@{}} World Knowledge \\ (5\%) \end{tabular}  &  \begin{tabular}{@{}L{11cm}@{}}
 \textsc{A:} May we see the wine list ?   \textsc{B:} Sure . Our special wine today is a 1989 Chardonnay .  \\  
 \textsc{A:} That sounds pretty good! How much is it ?   \textsc{B:} It's $\$4.25$ cents by the glass . The whole bottle is $\$22.25$ .  \\
 \textsc{A:} \colorbox{hl}{\textbf{I'd like a bottle}} please . \textsc{B:} I'll need to \colorbox{hl}{\textbf{see your ID}} please .  \\
 \textsc{A:} Here you go .   B: Sorry about the inconvenience, I had make sure you are over \rule{1cm}{1pt}.   
 \end{tabular} & \begin{tabular}{@{}L{2cm}@{}} \greencheck\, 21 years old \\ \redcross\, 30 years old \\\redcross\, 4 years old \\ \greencheck\, 18 years old  \end{tabular}                \\
\midrule
 \begin{tabular}{@{}L{2cm}@{}} Comparison \\ (24\%) \end{tabular}  &  \begin{tabular}{@{}L{11cm}@{}}
  \textsc{A:} Yes , sir. May I help you?   \textsc{B:} Please I'd like a ticket to New York.    \\  
  \textsc{A:} For today?   \textsc{B:} No, \colorbox{hl}{\textbf{early Saturday morning}}.  \\
 \textsc{A:} We have a flight that we'll put you there at \rule{1cm}{1pt}. Is that ok? \textsc{B:} \colorbox{hl}{\textbf{Nothing earlier?  I prefer flight at 9 thirty.}} \\\textsc{A:} I'm afraid not , unless you want a night flight. \textsc{B:} No, exactly not.
 \end{tabular} & \begin{tabular}{@{}L{2cm}@{}} \greencheck\,ten AM\\ \redcross\, 9:30 PM\\ \greencheck\,eleven AM \\ \redcross\, four AM \end{tabular}                \\
\midrule
\begin{tabular}{@{}L{2cm}@{}} Arithmetic \\ (5\%) \end{tabular}  &  \begin{tabular}{@{}L{11cm}@{}}
 \textsc{A:} How long do you want the house ? All summer ? \textsc{B:} No , just for six weeks.  \\ 
 \textsc{A:} I'm afraid I can only \colorbox{hl}{\textbf{rent it for two months}}. \\ \textsc{B:} My holiday is only \rule{1cm}{1pt}, but I think my brother and his family would take it for the \colorbox{hl}{\textbf{other two weeks}}. \\
 \end{tabular} & \begin{tabular}{@{}L{2cm}@{}}   \redcross\, six decades \\ \greencheck\, 45 days \\ \greencheck\, six weeks \\ \redcross\, two months 
 \end{tabular}                \\
\midrule
\begin{tabular}{@{}L{2cm}@{}} General Commonsense \\ (60\%) \end{tabular}  &  \begin{tabular}{@{}L{11cm}@{}}
 \textsc{A:} Do you get up early every morning ? \textsc{B:} About 6 in the morning.  \colorbox{hl}{\textbf{I like to walk to the office}}.  \\  
 \textsc{A:} Good habit. How long does it take ? \textsc{B:} \rule{1cm}{1pt}. Do you live alone ?  \\
 \textsc{A:} No , my little sister lives with me \ldots
 \end{tabular} & \begin{tabular}{@{}L{2cm}@{}} \greencheck\,20 minutes \\ \redcross\, 10 seconds \\ \greencheck\,15 minutes \\ \redcross\, 20 hours \end{tabular}                \\
 \midrule
 \begin{tabular}{@{}L{2cm}@{}} Others \\ (6\%) \end{tabular}  &  \begin{tabular}{@{}L{11cm}@{}}
 \textsc{A:} How long does a facial service take? \textsc{B:} We have half-hour and one-hour treatments.  \\  
 \textsc{A:} What's the regular price?  \textsc{B:} Well , \colorbox{hl}{\textbf{the half-hour}} facial costs $\$50$ and \colorbox{hl}{\textbf{the one-hour}} costs $\$80$. \\
 \textsc{A:} Good , I will take \rule{1cm}{1pt} facial. \textsc{B:} That's fine , madam.
 \end{tabular} & \begin{tabular}{@{}L{2cm}@{}} \greencheck\, the one hour\\ \redcross\, the 20 hour \\ \redcross\, the 80 second \\ \greencheck\, the half hour \end{tabular}               \\
\bottomrule
\end{tabular}
\caption{
Example dialogs and answer options from the \textsc{TimeDial} dataset, categorized by the nature of reasoning required to correctly answer them, along with the percentage of each reasoning category in the set of 100 sampled examples. The relevant key information in the dialog context is highlighted.
}
\label{tab:categories}
\vspace{-1em}
\end{table*}

\section{Dataset: \textsc{TimeDial}}\label{sec:data}

The \textsc{TimeDial} dataset is derived from DailyDialog data~\cite{li2017dailydialog}, which is a multi-turn dialog corpus containing over $13$K English dialogs. 
Dialogs in this dataset consist of turn-taking between two people on topics over $10$ broad categories, ranging from daily lives to financial topics.

\subsection{Data Collection}

Our data collection process involves two steps: (1) identifying dialogs that are rich in temporal expressions, and (2) asking human annotators to provide correct and incorrect options for cloze instances derived from these dialogs. We now describe these steps in detail. 

\paragraph{Temporal expression identification.}
Here, we select dialogs that are rich with temporal information, in order to focus on complex temporal reasoning that arises in natural dialogs. 
Temporal expressions are automatically identified with SUTime~\cite{chang2012sutime}, an off-the-shelf temporal expression detector.\footnote{\url{https://nlp.stanford.edu/software/sutime.shtml}} 
We keep only the dialogs with more than 3 temporal expressions and at least one expression that contains \textbf{numerals} like ``\textit{two weeks}'' (as opposed to non-numeric spans, like ``\textit{summer}'', ``\textit{right now}'', and ``\textit{later}''). 
In our initial experiment, we observe that language models can often correctly predict these non-numerical temporal phrases.

We note that temporal expressions containing numerals serve as more challenging sets of options than non-numerical ones. This filtering step results in 1,127 unique dialogs for further processing.

\paragraph{Human annotated options.}
Next, we make spans in the dialogs. For a dialog, we mask out each temporal expression that contains numerals, each resulting in a cloze question that is then sent for human annotation.

This resulted in 1,526 instances for annotation.
For each masked span in each dialog, we obtain human annotation to derive a fixed set of correct and incorrect options given the context. Concretely, given a masked dialog and a seed correct answer (i.e., the original text) for the masked span, the annotators\footnote{who are English linguists.} were asked to (1) come up with \textit{an alternative correct answer} that makes sense in the dialog adhering to commonsense, and (2) formulate \textit{two incorrect answers} that have no possibility of making sense in the dialog context. We highlight all time expressions in the context to make it easier for annotators to select reasonable time expressions. 

To ensure that the annotated incorrect options are not too trivially distinguishable by the models (as discussed in \S\ref{sec:task}), we define three rules for the annotators to follow. 
\begin{itemize}
\setlength\itemsep{-0.5em}
\item \textbf{Rule 1: Phrase Matching.} The rater should first try to pick another temporal span from the dialog context that makes syntactic/semantic sense (e.g., when the span is of the appropriate type, such as duration, for the masked span) but is still incorrect according to commonsense. 
\item \textbf{Rule 2: Numeral Matching.} If Rule 1 does not apply, raters should follow a relaxed version of Rule 1, whereby the incorrect option should contain any numeral occurring in the dialog context. 
\item \textbf{Rule 3: Open-ended.} If neither of the above rules is applicable, then raters can come up with an incorrect option using their own judgment. The two incorrect options are required to differ from each other as much as possible.
\end{itemize}
\noindent Rules-1\&2 are designed to confuse models that rely on shallow pattern matching. Finally, to ensure the quality of the human-annotated options, we perform a subsequent round of human validation on the gathered data. The validators identify and fix issues such as duplicate options, unreasonable or obscure annotations w.r.t natural usage, or ungrammatical annotations that do not fit in the context.

\subsection{Properties of \textsc{TimeDial}}

\begin{table}[!t]
\small
    \centering
    \begin{tabular}{l l}
    \toprule
    \# Dialog instances   & $1,104$  \\
    \# Temporal Expressions &  $1,985$  \\
    \# Avg. Turns Per Dialog & $11.7$ \\
    \# Avg. Words Per Turn & $16.5$\\
    \# Avg. Time Spans Per Dialog & $3.0$ \\
    \midrule
    \multicolumn{2}{c}{\emph{Incorrect Options}} \\
    \midrule
    \% Phrase Matching & $16.3$ \% \\
    \% Numeral Matching &  $49.6$ \% \\
    \% Open-ended &  $45.4$ \% \\
    \bottomrule
    \end{tabular}
    \caption{Statistics of our \textsc{TimeDial} challenge set.} 
    \label{tab:stats}
    \vspace{-1em}
\end{table}

Table~\ref{tab:stats} shows statistics of \textsc{TimeDial}. 
The dataset contains over 1.1K test instances.
Each dialog contains 11.7 turns and 3 temporal expressions on average, presenting richer and more complex context compared to the recent single-sentence-based temporal question answering benchmarks \cite[e.g.,][]{zhou-etal-2019-going,vashishtha-etal-2020-temporal}. As above, each test instance contains two correct answers and two incorrect ones.\footnote{We also collected 342 extra instances for which the annotators deem there is only one unique correct answer for the context. Thus, each of those instances contains one correct option and two incorrect ones. We release those instances along with the dataset, though we did not include them in empirical study in this paper.} 
Over half of the incorrect options are annotated based on phrase and numeral matching from context, which pose a significant challenge for models relying on shallow text matching, as we show in our experimental analysis (\S\ref{sec:exp}).

Answering different instances in the dataset requires different types of core reasoning abilities, such as comparison, arithmetic inference, or reasoning based on world knowledge or general commonsense. To facilitate fine-grained analysis, we also annotate the \textbf{reasoning categories} for a randomly sampled set of 100 dialogs. Though each instance can involve multiple reasoning types, we associate it with one predefined category label that indicates the primary type of reasoning it requires. Table~\ref{tab:categories} shows the category distribution and examples in each of the category. 
We observe that the dataset requires general commonsense for $60\%$ of the dialogs, making it the most common reasoning type.

\section{Modeling}\label{sec:models}

\begin{figure*}[!t]
\centering
\includegraphics[width=0.95\textwidth]{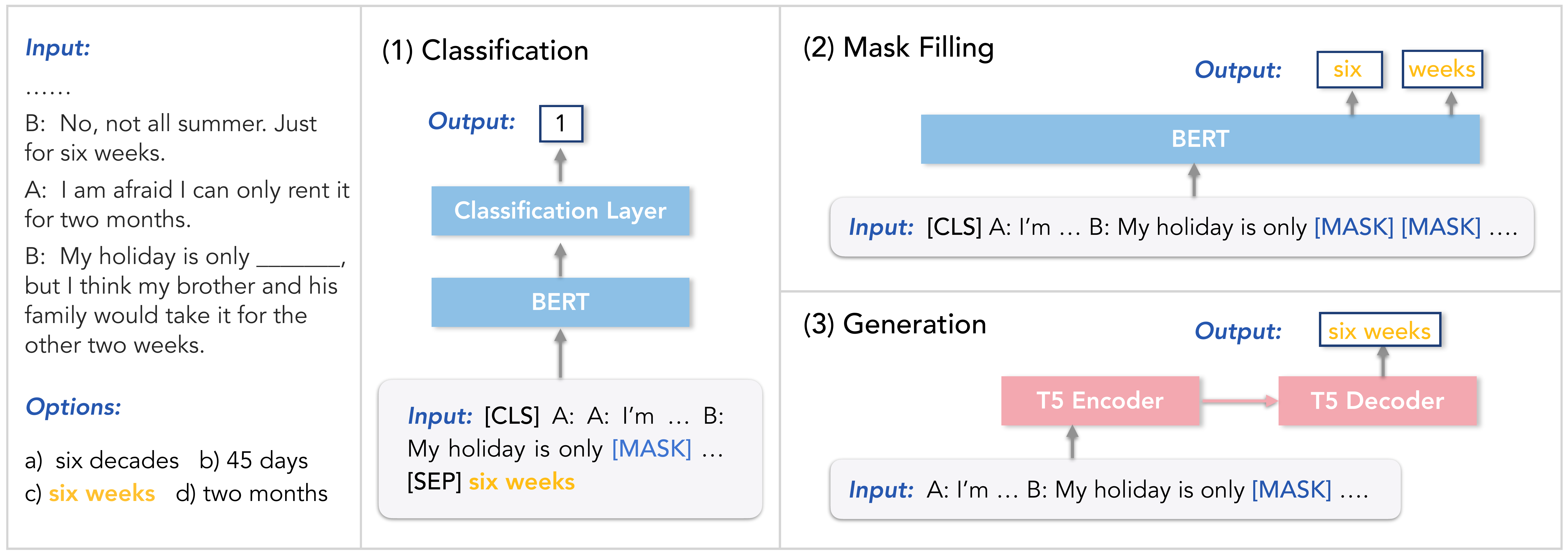}
\caption{
We study three modeling paradigms for the task, based on BERT and T5, including (1) Classification, (2) Mask Filling, and (3) Generation (\S\ref{sec:model-paradigm}). The models are finetuned with various training data, as discussed in \S\ref{sec:training}.
}
\label{fig:model} 
\end{figure*}

We consider a broad set of methods and evaluate their performance on our challenge \textsc{TimeDial} dataset. These methods vary in terms of the modeling paradigms, the scope of the dialog contexts, and training settings. In particular, they encompass the major ways pre-trained LMs are currently used in downstream tasks (\S\ref{sec:model-paradigm}) which often outperform earlier specialized non-pretrained models. 
We also consider different lengths of context used in reasoning, varying by their vicinity to the masked span (\S\ref{sec:dialog-context}). Finally, we study different training settings, including zero-shot, in-domain, and out-of-domain training (\S\ref{sec:training}). 

\subsection{Modeling Paradigms}\label{sec:model-paradigm}
We experiment across three major modeling paradigms: (i) Binary Classification, (ii) Mask Filling, and (iii) Generation. Figure~\ref{fig:model} shows the different architectures. For each test instance, the model takes as input a pair of (masked dialog context, candidate), and outputs a score measuring how likely the candidate being a correct answer. Based on the prediction scores of all options, the model then chooses the top two positive candidates as the predicted answer for the instance.
Each paradigm of models is finetuned using training data from different domains, as discussed in \S\ref{sec:training}. 

\subsubsection{Binary Classification}
In this setting, we formulate the task as a binary classification problem, i.e., we use a classifier to measure the probability of the candidate in the (masked dialog context, candidate) pair being a correct answer.
Any powerful LM ---  e.g., \textsc{BERT}~\cite{devlin-etal-2019-bert}, \textsc{ALBERT}~\cite{lan2019albert}, \textsc{RoBERTa}~\cite{liu2019roberta}, etc. can be used to build the classifier.

This method's key challenge is the lack of annotated training data for direct supervision. We generate weak supervision training data as follows. In an unlabeled corpus, we use the SUTime tool to annotate temporal spans. We mask each temporal span in this corpus and use the masked text as one positive example for binary classification. To generate negative example, we randomly sample another temporal span from the dialog context and use it as a negative example for the masked temporal span.
The resulting data is noisy because the randomly sampled temporal span can also logically fit in the masked span in the given context; however, we assume the likelihood of that happening is low. We leave drawing harder negative instances using heuristics to future work.
\subsubsection{Mask Filling}

We also use the mask filling approach of \textsc{BERT}-like mask language models (MLMs). For each dialog context and a candidate temporal span of $m$ tokens, we replace the blank in the dialog context with $m$ masked tokens. We then evaluate the likelihood of predicting the temporal span tokens for those masked positions, and make average across the positions. A key advantage of this method is that we can directly apply a BERT model in the \textit{zero-shot} manner since the model was pre-trained in the same way, as for accommodating for \texttt{[MASK]} fillings. Additionally, we also finetune \textsc{BERT}'s MLM for learning task specific properties. 

\subsubsection{Generation}

The third method is a fully generative approach using the \emph{text-to-text} paradigm of T5~\cite{2020t5}. Given a masked dialog context, the model is trained to generate the masked text in an encoder-decoder framework. As a result, evaluating the likelihood of generating the given temporal span (normalized with the length of the span) is used as the probability of it being correct. 
Similar to mask filling, we use \textsc{T5} either in a zero-shot manner or with additional fine-tuning.

\subsection{Dialog Context}\label{sec:dialog-context}
We aim to study the influence of context on a model's temporal reasoning in dialog by incorporating varying scopes of dialog context based on their vicinity to the target span. Since the dialogs in \textsc{TimeDial} are rich in temporal concepts, we want to evaluate LMs' dependence on shallow text matching vs. the ability to accurately understand the causal relations between those concepts (see Table~\ref{tab:spurious-feature-result}). We use the following three settings:

\begin{itemize}
\setlength\itemsep{-0.5em}
    \item \textbf{Full} context, where the model is presented with the complete available dialog to reason on. Due to our design of challenging negatives, the full context can often confuse models that rely on shallow cues.
    \item \textbf{Local} context, where we provide only with the  utterances that immediately precede and follow the target utterance.
    \item \textbf{Target} context, where the context is restricted to only the particular utterance that contains the masked span.

\end{itemize}
\begin{table}[t]
\centering
\small
\begin{tabular}{lccc}
\toprule
\rowcolor[gray]{0.97} \multicolumn{4}{c}{\emph{Mask Filling} and \emph{Generation}} \\ \cmidrule{1-4}
              & \#~Train        &  \#~Dev          &                \\ \cmidrule{2-3}
In-domain (Daily)         & 14.5K        & 2.4K         &                \\
Out-domain (Meena)         & 1.26M        & 23K          &                \\ \midrule
\rowcolor[gray]{0.97} \multicolumn{4}{c}{\emph{Classification}}        \\ \cmidrule{1-4}
              & \#~Train        &  \#~Dev          & \#~Spans        \\ \cmidrule{2-4}
In-domain (Daily)        & 58.0K          & 9.6K        & 2,153          \\
Out-domain (Meena)          & 5.04M        & 92K         & 38,750         \\ 
\bottomrule
\end{tabular}
\caption{
Number of training and development instances for different settings. An instance is derived by masking one temporal span of a dialog. For classification, we draw 3 negative samples per positive sample.
``\#~Spans'' is the size of temporal span pool from which negative samples are drawn for weak supervision.
}
\label{tab:training-data}
\vspace{-1em}
\end{table}

\subsection{Training Details}
\label{sec:training}

For all models, we consider two common training settings, e.g., in-domain data, which is typically small, and out-of-domain training where a large amount of data is available. Table~\ref{tab:training-data} shows training data statistics. For mask-filling and generation, we also evaluate in a zero-shot setup with no finetuning. 

\paragraph{In-domain training.}
Our challenge \textsc{TimeDial} test set is derived from contextually rich dialogs from the DailyDialog dataset, based on the number of temporal spans. However, this still leaves remaining data with less than 3 temporal spans or with no numeric span. By masking each temporal span in each dialog, we obtain 14.5K training instances to use in our domain specific fine-tuning. 

\paragraph{Out-of-domain training.}
In this setting, we consider a much larger corpus from a general domain. Specifically, we use the large scale training set based on the Meena dataset \citet{adiwardana2020towards}, which is mined and filtered from public domain social media conversations over 341GB of text (40B words).\footnote{We acquired a trimmed down version of the Meena dataset by contacting the authors.} Compared to the above in-domain data from DailyDialog which were manually written by human annotators in a clean and consistent way, the dialogs in the Meena corpus tend to be noisy, casual, and usually short.
Like our DailyDialog processing, we identify all temporal expressions for dialogs in Meena using SUTime.

\section{Experiments and Analyses}\label{sec:exp}

\begin{table}[t]
\small
\centering
\begin{tabular}{L{3cm}R{2cm}}
\toprule
\textsc{Size-Train}  & \bf \textit{$2$-best Acc (\%)}  \\
\midrule
\rowcolor[gray]{0.97} \multicolumn{2}{c}{\emph{Classification (BERT)}} \\
\midrule
\textsc{Base-Out}   & $43.1$       \\
\textsc{Base-In}   & $51.1$      \\
\textsc{Large-Out}    & $48.7$       \\
\textsc{Large-In}   & $53.2$       \\
\midrule
\rowcolor[gray]{0.97} \multicolumn{2}{c}{\emph{Mask Filling (BERT)}} \\
\midrule
\textsc{Base-Zero}  & $44.8$      \\
\textsc{Base-Out}   & $47.4$      \\
\textsc{Base-In}   & $67.4$       \\
\textsc{Large-Zero} & $47.7$    \\
\textsc{Large-Out}  & $54.8$       \\
\textsc{Large-In}  & $70.0$      \\
\midrule
\rowcolor[gray]{0.97} \multicolumn{2}{c}{\emph{Generation (T5)}} \\
\midrule
\textsc{Base-Zero}         & $39.8  $     \\
\textsc{Base-Out}          & $50.6  $     \\
\textsc{Base-In}          & $59.2 $      \\
\textsc{Large-Zero}         & $39.1$     \\
\textsc{Large-Out}         & $61.9$      \\ 
\textsc{Large-In}         & $\mathbf{74.8}$      \\
\midrule
\bf Human & $\mathbf{97.8}$   \\
\bottomrule
\end{tabular}
\caption{
Model and human performance on \textsc{TimeDial}. 
\textsc{Base} and \textsc{Large} denote the size of the pre-trained BERT and T5; 
\textsc{Zero}, \textsc{In}, and \textsc{Out} denote that the model is zero-shot (with no finetuning), fintuned using the in-domain DailyDialog data, or finetuned using the out-of-domain Meena data, respectively. THe full dialog context is used for all models. 
}
\label{tab:model-perform}
\vspace{-1em}
\end{table}

Using the proposed \textsc{TimeDial} challenge set, we next conduct extensive experiments and analyses on the different model variants and context settings. We use either 4x4 or 8x8 Cloud TPUs V3 pod slices\footnote{\url{https://cloud.google.com/tpu}} for fine-tuning and one V100 GPU for inference. We provide more details of the experiment configurations in the appendix.

\paragraph{Evaluation.}
Since each example of \textsc{TimeDial} contains two correct answers, we
report the metric \textit{$2$-best accuracy}, which measures whether \textit{both} of the model's top-ranked 
answers are correct. In other words, if the model erroneously ranks an incorrect answer over a correct one, we consider it to be an error case. Note that we use the ranking-based metric as opposed to classification-based ones (for example, by asking the model to classify whether each individual candidate answer is correct or not \citep[e.g.,][]{zhou-etal-2019-going}) and because it presents a stricter measure that penalizes any incorrect answers being ranked over correct answers, and the ranking metric is not influenced by specific choices of the threshold hyperparameter that cuts off positive and negative predictions.

\subsection{Model Performance}

\begin{table*}[!t]
\centering
\footnotesize
\begin{tabular}{L{8cm}|C{2cm}|C{0.7cm}|C{0.7cm}|C{0.7cm}|C{0.7cm}}
\bf Dialog Context   & \bf Options & \sc Gold & \sc CLS & \sc MF & \sc GEN   \\
\toprule

\begin{tabular}{@{}L{8cm}@{}}
\textsc{A:}  What's the date \textbf{today}? \\
\textsc{B:} \textbf{Today} is \textbf{September 28th, 2007}. \\
\textsc{A:} I have a meeting this \textbf{afternoon}.   \\
\textsc{B:} When will it begin? \\  
\textsc{A:} It will begin at \textbf{three o'clock}. What's the time \textbf{now}?   \\
\textsc{B:} It is \rule{1cm}{1pt}. \\  
\textsc{A:} I have to go \textbf{now}. I don't want to be late.   \\
\textsc{B:} Don't worry, time is enough. \\  
 \end{tabular} & 
 \begin{tabular}{@{}C{2cm}@{}}

half past one  \\[.2cm] quarter to two \\[.2cm] half past three \\[.2cm] half past nine
 \end{tabular} &

 \begin{tabular}{@{}C{0.7cm}@{}}
 \greencheck\\[.2cm] \greencheck\\[.2cm] \redcross \\[.2cm]\redcross
 \end{tabular} & 
 \begin{tabular}{@{}C{0.7cm}@{}} 
 \greencheck\\[.2cm]\redcross \\[.2cm] \greencheck\\[.2cm]  \redcross
 \end{tabular} & 
 \begin{tabular}{@{}C{0.7cm}@{}}
 \redcross \\[.2cm]\redcross \\[.2cm] \greencheck\\[.2cm] \greencheck 
 \end{tabular} & 
 \begin{tabular}{@{}C{0.7cm}@{}} 
 \greencheck \\[.2cm]\redcross \\[.2cm] \greencheck\\[.2cm]  \redcross
 \end{tabular}   \\
\midrule
\begin{tabular}{@{}L{8cm}@{}}
\textsc{A:} Doctor, I feel much better \textbf{now}. Will I be able to go home \textbf{some time this week}?\\   
\textsc{B:} That's good to hear. You've had an ideal recovery from your operation. We're going to send you home tomorrow.  \\ 
\textsc{A:} Do you think I can get back to work \textbf{very soon}? \\
\textsc{B:}Don't be in such a hurry. I'm confident that you'll be completely recovered in \rule{1cm}{1pt}. \\
\textsc{A:} Is there anything I should do? \\
\textsc{B:} You'd better have a good rest for \textbf{a week}. 
\end{tabular}       & \begin{tabular}{@{}C{2cm}@{}}4 to 6 weeks \\[.2cm]5 to 7 weeks \\[.2cm]a week \\[.2cm]a day  \end{tabular}         & \begin{tabular}{@{}C{0.7cm}@{}}  \greencheck\\[.2cm] \greencheck\\[.2cm] \redcross \\[.2cm]\redcross\end{tabular}              &    \begin{tabular}{@{}C{0.7cm}@{}} \redcross \\[.2cm]\greencheck \\[.2cm] \greencheck\\[.2cm] \redcross \end{tabular} &  \begin{tabular}{@{}C{0.7cm}@{}} \redcross \\[.2cm]\redcross \\[.2cm] \greencheck\\[.2cm] \greencheck \end{tabular} & \begin{tabular}{@{}C{0.7cm}@{}} \greencheck \\[.2cm]\redcross \\[.2cm] \greencheck\\[.2cm] \redcross \end{tabular}   \\
\bottomrule
\end{tabular}
\vspace{-2pt}
\caption{Example prediction errors made by different models for cases with challenging options, based on the phrase and numeral matching rules (\S\ref{sec:data}). \textsc{Gold} denotes the true labels.
The model predictions show that the models get confused by learning shallow text matching in terms of pre-existing temporal concepts (marked by bold faced text) in the context.}
\label{tab:spurious-feature-result}
\vspace{-1em}
\end{table*}

Table~\ref{tab:model-perform} shows model results and human performance. Human performance achieves a near-perfect level ($97.80$, with Cohen's kappa score of 0.86 showing almost perfect inter-rater agreement \cite{landis1977measurement}). 

\paragraph{Overall.} 
The generation model based on \textsc{T5-Large} and finetuned on the in-domain DailyDialog data achieves the best performance. However, its $2$-best accuracy ($74.8$) lagged far behind the human performance, demonstrating the difficulty of the \textsc{TimeDial} challenge set.

\paragraph{Zero-shot vs. out-of-domain vs. in-domain.} 
When comparing the different training data setup, we observe that models with in-domain training using the DailyDialog data (e.g., \textsc{Large-In}) consistently outperforms those trained on the large out-of-domain Meena dataset (e.g., \textsc{Large-Out}). Both setups outperform the zero-shot models (without any fine-tuning) (e.g., \textsc{Large-Zero}). 
The results show that the large LMs still highly depend on in-domain or at least dialog data to grasp and enhance their temporal reasoning ability in dialog context. Further, we see increasing performance with increasing model size, which is not unexpected given the complexity of the task.

\subsection{Error Analysis}

Next, we analyze the different types of errors based on different rules for negative option creation in the annotation process. In particular, the \emph{phrase matching} rule picks an exact time span from the dialog context, and \emph{numeral matching} picks numerals from the dialog context. Thus, models picking those incorrect options imply reliance on spurious shallow text matching features.

\begin{figure}[!t]
\centering
\includegraphics[width=0.42\textwidth]{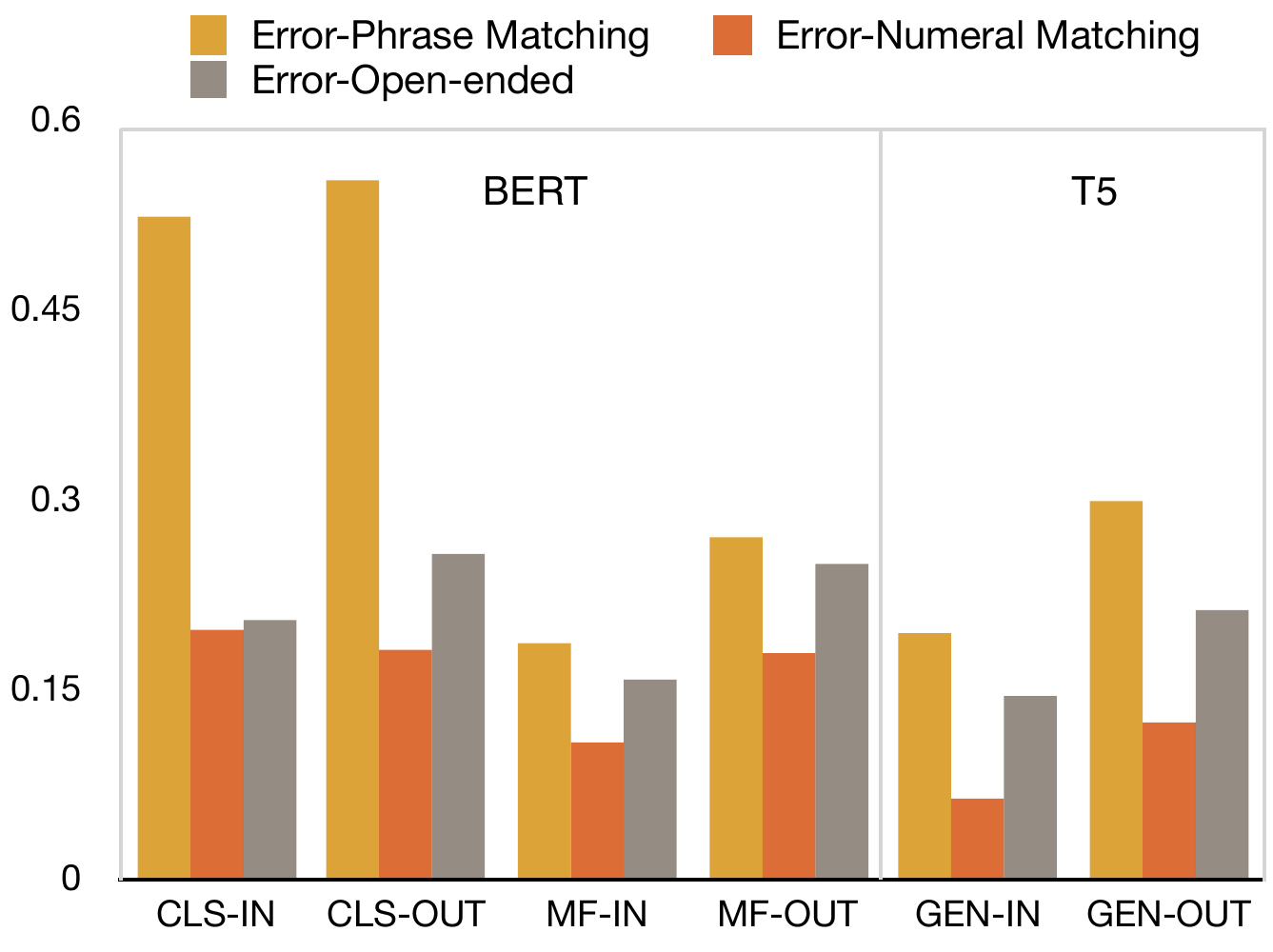}
\vspace{-7pt}
\caption{
Percentage of errors on options created by different rules.
\textsc{CLS, MF}, and \textsc{GEN} represent classification, mask-filling, and generation models, respectively; and \textsc{In} and \textsc{Out} denote in-domain and out-of-domain training. All models are of large size.
}
\label{fig:error-types} 
\vspace{-1em}
\end{figure}

Figure~\ref{fig:error-types} shows the percentage of errors in terms of the different rules. For example, the BERT-based classification model \textsc{CLS-In} erroneously picks 52\% of negative options created by the phrase matching rule as correct answers (i.e., by ranking those negative options over the true correct options). We observe that the various models are all most vulnerable to the phrase matching options compared to other types of negative options, showing that they rely on spurious text matching to a significant extent.
Between \textsc{BERT} and \textsc{T5}, we find \textsc{T5} being more robust to shallow text matching.

\begin{table}[!t]
\centering
\small
\begin{tabular}{lrrrr}
\toprule
{\bf Size} & \multicolumn{2}{c}{{\sc BASE}}  & \multicolumn{2}{c}{{\sc LARGE}} \\
{\bf Training}      & \textsc{In} & \textsc{Out} & \textsc{In} & \textsc{Out} \\
\cmidrule{1-5}
\rowcolor[gray]{0.97} \multicolumn{5}{c}{\emph{Classification (BERT)}} \\
\cmidrule{1-5}
\sc Target & 50.5     & 40.0      & 50.5     & 47.5      \\
\sc Local  & \cellcolor[gray]{0.9}$+$\textbf{ 3.4}   & \cellcolor[gray]{0.9}\textbf{$+$ 3.3}      & \cellcolor[gray]{0.9}\textbf{$+$ 7.5}     & \cellcolor[gray]{0.9}\textbf{$+$ 2.0}      \\
\sc Full   & $-$ 0.6     & $-$ 0.1      & $+$ 2.7     & $+$ 1.2      \\ \cmidrule{1-5}
\rowcolor[gray]{0.97} \multicolumn{5}{c}{\emph{Mask Filling (BERT)}} \\
\cmidrule{1-5}               
\sc Target & 57.8     & 44.3      & 60.3       & 46.8      \\
\sc Local  & $+$ 5.4     & $+$ 3.0     & $+$ 8.1     & $+$ 4.9      \\
\sc Full   & $+$ 9.6     & $+$ 3.1      & $+$ 9.6     & $+$ 8.0      \\ 
\cmidrule{1-5}
\rowcolor[gray]{0.97} \multicolumn{5}{c}{\emph{Generation (T5)}} \\
\cmidrule{1-5}                             
\sc Target & 55.5     & 45.9      & 66.7     & 56.1      \\
\sc Local  & $+$ 3.7    & $+$ 2.7      & $+$ 6.1     & $+$ 3.7      \\
\sc Full   & $+$ 3.7     & $+$ 4.7      & $+$ 8.2     & $+$ 5.8     \\
\bottomrule
\end{tabular}
\caption{
Impact of dialog context on reasoning accuracy. \textsc{In} and \textsc{Out} denote in-domain and out-of-domain training, respectively.
We use \emph{2-best accuracy} of \emph{target} context as reference and report the absolute changes in performance of \emph{local} and \emph{full} context, respectively. 
Local dialog context results in better performance to full dialog context on $5$ of the $12$ cases, which are highlighted in the table.}
\label{tab:results-context}
\end{table}

Table~\ref{tab:spurious-feature-result} provides further examples of prediction errors, illustrating confusions due to shallow text matching. 
In the first dialog, both incorrect answers already partially occur in the context or are related to preexisting concepts (i.e., \emph{``three''} to \emph{``three o'clock''}, and \emph{``nine''} to \emph{``September''}).
All the three models were confused and chose either of the two as the top prediction for the blank, even though the options clearly violate the context.
Interestingly, the mask filling model was completely confused and ranked both incorrect answers over the correct ones. Similarly in the second example, the models fail to capture the contextual semantics.

\subsection{Influence of Dialog Context}

Table~\ref{tab:results-context} shows how different scopes of dialog context (\S\ref{sec:dialog-context}) affect model performance. First, the most restrictive target-only context is insufficient for accurate reasoning, by producing the weakest performance of most models. This highlights the importance of context information for temporal commonsense reasoning in dialog, which differs from previous temporal reasoning studies based on limited context (e.g., single-sentence question answering).
Second, we note that the full dialog context does not always lead to the best performance. In $5$ out of the $12$ cases, using the local context yields equal or higher reasoning accuracy. The results show that the LMs still fall short of properly modeling the rich dialog contexts and making effective use of all information to do reasoning.

\begin{figure}[!t]
\centering
\includegraphics[width=0.42\textwidth]{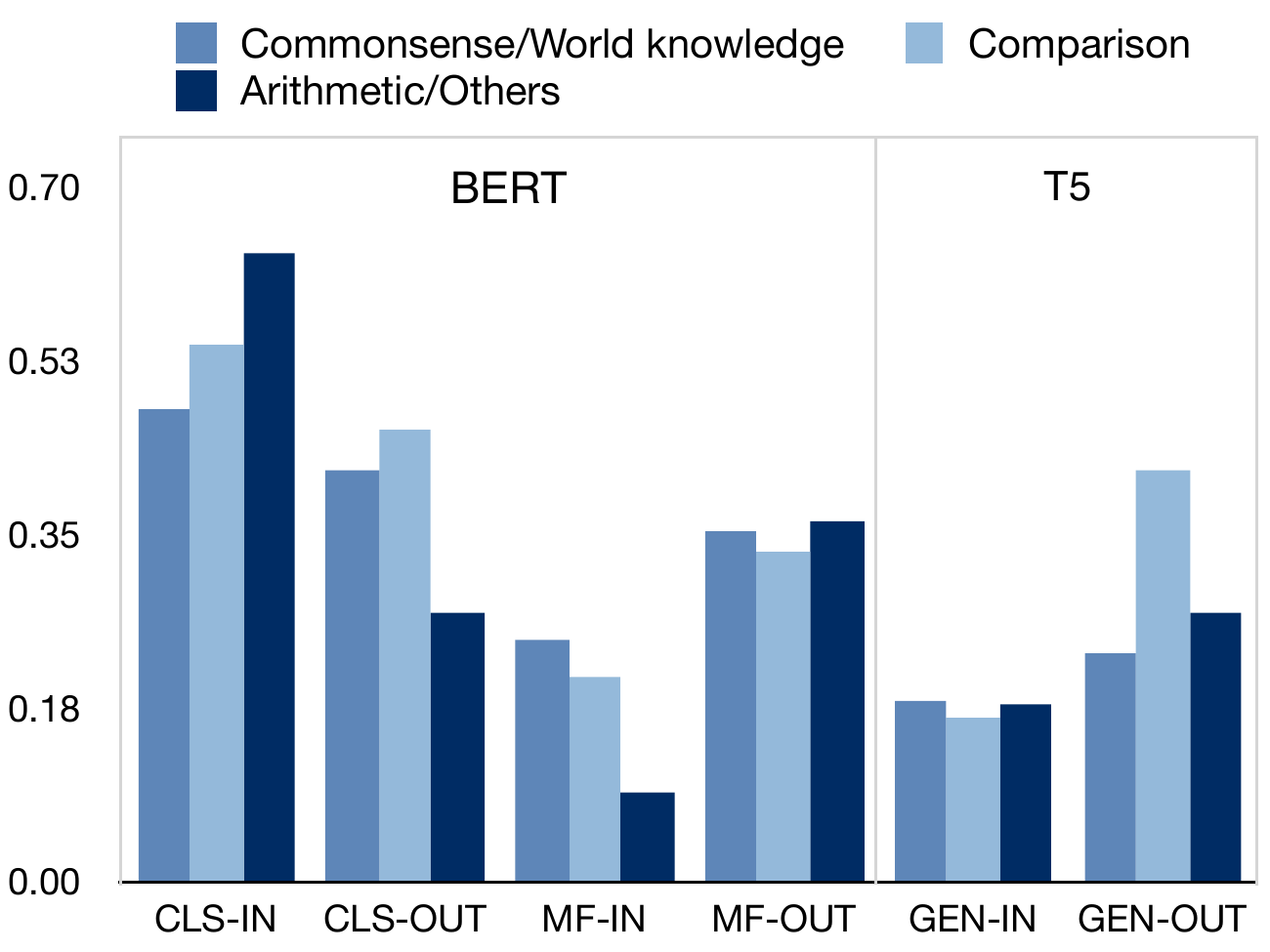}
\vspace{-7pt}
\caption{
Percentage of errors on different reasoning types. \textsc{CLS, MF}, and \textsc{GEN} represent classification, mask-filling, and generation models, respectively. All models are of large size.
}
\label{fig:error-category} 
\vspace{-1em}
\end{figure}

\subsection{Errors of Reasoning Categories}

Figure~\ref{fig:error-category} shows the percentage of errors in each reasoning category.  
We observe that the models tend to make non-trivial portions of errors on commonsense/world knowledge questions. For example, the strongest model, \textsc{T5 Gen-In}, failed on 18\% of the instances that require commonsense or world knowledge, while \textsc{BERT CLS-In} made errors on 48\% of such instances. The performance on comparison-based instances seems similar.

\section{Related Work}\label{sec:related}
\paragraph{Temporal commonsense reasoning.}
Early studies related to temporal analysis define time in the context of sets and relations~\cite{bruce1972model, allen1983maintaining}. 
More recent works often associate time with events and focus on identifying time expressions~\cite{chang2012sutime,angeli2012parsing, lee2014context}, extracting temporal relations among events ~\cite{setzer2000annotating, pustejovsky2005temporal, lapata2006learning, chambers2007classifying,ning-etal-2018-improving}, and timeline construction \cite{do-etal-2012-joint, leeuwenberg2018temporal}.

Some recent work has focused on building challenging benchmarks for temporal commonsense reasoning. 
Story Cloze Test focuses on stereotypical causal temporal and causal relations between events~\cite{mostafazadeh-etal-2016-corpus}.
\citet{vashishtha-etal-2020-temporal} recast temporal reasoning datasets for event duration and event ordering into the natural language inference (NLI) format.
Turque \cite{ning-etal-2020-torque} is an reading comprehension dataset where the model needs to answer questions such as ``what happens before/after [event]''. 
Most related to our work is McTaco \cite{zhou-etal-2019-going}, a dataset for evaluating temporal commonsense in the form of multiple-choice reading comprehension, where the context usually consists of a single sentence. Our work instead studies temporal commonsense reasoning in dialogs which often require significant commonsense and world knowledge to reason over rich context \citep{qin2019conversing,dinan2018wizard}.

\paragraph{Commonsense reasoning with LMs.}
With the recent success of large pre-trained language models (LMs)~\cite{devlin-etal-2019-bert, brown2020language}, it is an open question whether these models, pretrained on large amounts of data, capture commonsense knowledge. 
Several works have been proposed to assess the ability of LMs for commonsense or numerical reasoning ~\cite{zhang-etal-2020-language-embeddings,Bouraoui2020InducingRK}, or to mine commonsense knowledge from LMs~\cite{davison-etal-2019-commonsense}. \citet{lin-etal-2020-birds} showed that state-of-the-art LMs such as BERT and RoBERTa performs poorly on numerical reasoning tasks without any finetuning.
Works have also been proposed to improve language model's commonsense reasoning \cite{qin2020backpropagation,qin2019counterfactual,zhou-etal-2020-temporal} and numerical reasoning abilities~\cite{geva2020injecting}.
In our work, we study several modeling approaches and finetuning settings of large LMs, and establish strong baselines for temporal commonsense reasoning in dialogs.

\section{Conclusions}\label{sec:conclusions}

We introduced \textsc{TimeDial}, a challenge set consistting of 1.1K multiple-choice cloze questions  for temporal commonsense reasoning in dialog. 
The dataset is carefully curated to evaluate a models' ability to do temporal commonsense/numerical reasoning over dialog context. 
In order to establish strong baselines and provide information on future model development, we conducted extensive experiments with state-of-the-art language models with different settings: the scope of context, weak supervision strategies, and learning objectives. 
While humans can easily answer these questions (97.8\% accuracy), even our best model variant (T5-large with in-domain training) struggles on this challenge set (73\%).
Moreover, our qualitative error analyses show that these large language models often rely on shallow, spurious features (particularly text matching) when answering these questions, instead of truly doing reasoning over the context.

\balance

\bibliography{acl2020}
\bibliographystyle{acl_natbib}

\clearpage

\appendix

\onecolumn

\section{Configurations}
We provide all model and training configurations used across our experiments:

\subsection{BERT Experiments for Classification and Mask-Filling}
\vspace{1em}
\begin{itemize}
\item Model configuration for \textsc{BERT-Base} classification and mask-filling:
\begin{small}
\begin{verbatim}
    attention_dropout_rate: 0.1
    dropout_rate: 0.1
    hidden_activation: gelu
    hidden_size: 768
    initializer_range: 0.02
    intermediate_size: 3072
    max_position_embeddings: 512
    num_attention_heads: 12
    num_layers: 12
    type_vocab_size: 2
    vocab_size: 30522
\end{verbatim}
\end{small}

\item  Model configuration for \textsc{BERT-Large} classification and mask-filling:
\begin{small}
\begin{verbatim}
    attention_dropout_rate: 0.1
    dropout_rate: 0.1
    hidden_activation: gelu
    hidden_size: 1024
    initializer_range: 0.02
    intermediate_size: 4096
    max_position_embeddings: 512
    num_attention_heads: 16
    num_layers: 24
    type_vocab_size: 2
    vocab_size: 30522
\end{verbatim}
\end{small}

\item Training configuration for classification with \textsc{BERT-Base} and in-domain data:
\begin{small}
\begin{verbatim}
    num_classes: 2
    train_data:
      global_batch_size: 128 
      seq_length: 512
    validation_data:
      global_batch_size: 32
      seq_length: 512
    trainer:
      max_to_keep: 3
      checkpoint_interval: 1000
      decay_steps: 30000
      end_learning_rate: 0.0
      initial_learning_rate: 1.0e-5
      power: 1.0
      optimizer: adam
      warmup_steps: 5000
      steps_per_loop: 1000
      train_steps: 30000
      validation_steps: 256
\end{verbatim}
\end{small}

\item Training configuration for classification with \textsc{BERT-Large} and in-domain data:
\begin{small}
\begin{verbatim}
    num_classes: 2
    train_data:
      global_batch_size: 128 
      seq_length: 512
    validation_data:
      global_batch_size: 32
      seq_length: 512
    trainer:
      max_to_keep: 3
      checkpoint_interval: 1000
      decay_steps: 100000
      end_learning_rate: 0.0
      initial_learning_rate: 1.0e-6
      power: 1.0
      optimizer: adam
      warmup_steps: 10000
      steps_per_loop: 1000
      train_steps: 100000
      validation_steps: 3000
\end{verbatim}
\end{small}

\item Training configuration for classification with \textsc{BERT-Base} and out-domain data:
\begin{small}
\begin{verbatim}
    num_classes: 2
    train_data:
      global_batch_size: 128 
      seq_length: 512
    validation_data:
      global_batch_size: 128
      seq_length: 512
    trainer:
      max_to_keep: 3
      checkpoint_interval: 5000
      decay_steps: 500000
      end_learning_rate: 0.0
      initial_learning_rate: 1.0e-6
      power: 1.0
      optimizer: adam
      warmup_steps: 10000
      steps_per_loop: 1000
      train_steps: 500000
      validation_steps: 512
\end{verbatim}
\end{small}


\item Training configuration for classification with \textsc{BERT-Large} and out-domain data:
\begin{small}
\begin{verbatim}
    num_classes: 2
    train_data:
      global_batch_size: 128 
      seq_length: 512
    validation_data:
      global_batch_size: 128
      seq_length: 512
    trainer:
      max_to_keep: 3
      checkpoint_interval: 5000
      decay_steps: 500000
      end_learning_rate: 0.0
      initial_learning_rate: 1.0e-6
      power: 1.0
      optimizer: adam
      warmup_steps: 10000
      steps_per_loop: 1000
      train_steps: 500000
      validation_steps: 512
\end{verbatim}
\end{small}

\item Training configuration for mask-filling with \textsc{BERT-Base} and in-domain data:
\begin{small}
\begin{verbatim}
    train_data:
      global_batch_size: 128 
      seq_length: 512
      max_predictions_per_seq: 20
    validation_data:
      global_batch_size: 128
      seq_length: 512
      max_predictions_per_seq: 20
    trainer:
      checkpoint_interval: 2000
      max_to_keep: 30
      decay_steps: 30000
      end_learning_rate: 0.0
      initial_learning_rate: 1.0e-8
      power: 1.0
      optimizer: adam
      warmup_steps: 5000
      steps_per_loop: 1000
      train_steps: 30000
      validation_interval: 1000
\end{verbatim}
\end{small}

\item Training configuration for mask-filling with \textsc{BERT-Large} and in-domain data:
\begin{small}
\begin{verbatim}
    train_data:
      global_batch_size: 128 
      seq_length: 512
      max_predictions_per_seq: 20
    validation_data:
      global_batch_size: 128
      seq_length: 512
      max_predictions_per_seq: 20
    trainer:
      checkpoint_interval: 2000
      max_to_keep: 30
      decay_steps: 30000
      end_learning_rate: 0.0
      initial_learning_rate: 1.0e-8
      power: 1.0
      optimizer: adam
      warmup_steps: 5000
      steps_per_loop: 1000
      train_steps: 30000
      validation_interval: 1000
\end{verbatim}
\end{small}

\item Training configuration for mask-filling with \textsc{BERT-Base} and out-domain data:
\begin{small}
\begin{verbatim}
    train_data:
      global_batch_size: 512 
      seq_length: 512
    validation_data:
      global_batch_size: 512
      seq_length: 512
    trainer:
      checkpoint_interval: 5000
      max_to_keep: 10
      decay_steps: 300000
      end_learning_rate: 0.0
      initial_learning_rate: 1.0e-6
      power: 1.0
      optimizer: adam
      warmup_steps: 10000
      steps_per_loop: 1000
      train_steps: 300000
      validation_steps: 1000
\end{verbatim}
\end{small}

\item Training configuration for mask-filling with \textsc{BERT-Large} and out-domain data:
\begin{small}
\begin{verbatim}
    train_data:
      global_batch_size: 512 
      seq_length: 512
    validation_data:
      global_batch_size: 512
      seq_length: 512
    trainer:
      checkpoint_interval: 5000
      max_to_keep: 10
      decay_steps: 300000
      end_learning_rate: 0.0
      initial_learning_rate: 1.0e-6
      power: 1.0
      optimizer: adam
      warmup_steps: 10000
      steps_per_loop: 1000
      train_steps: 300000
      validation_steps: 1000
\end{verbatim}
\end{small}
\end{itemize}

 \subsection{T5 Experiments for Generation}
 
\begin{itemize}
\vspace{1em}

\item The training configuration for generation with \textsc{T5-Base} and in-domain data:
\begin{small}
\begin{verbatim}
  encoder_seq_length: 512
  decoder_max_length: 128
  train_batch_size: 128
  max_train_steps: 100000
  valid_batch_size: 128
  dropout_rate: 0.2
  optimizer: adam
  learning_rate: 1.0e-6
\end{verbatim}
\end{small}

\item The training configurations for generation with \textsc{T5-Base}/\textsc{Large} and in-domain/out-domain data are similar as above, except that the \texttt{learning rate} is set to \texttt{5.0e-6} for \textsc{T5-Large} in-domain data,  \texttt{5.0e-4} for \textsc{T5-Base} out-domain data, and \texttt{1.0e-4} for \textsc{T5-Large} out-domain data.

\end{itemize}

\end{document}